\documentclass[12pt]{article}

\usepackage[utf8]{inputenc}
\usepackage[T1]{fontenc}
\usepackage{fullpage}
\usepackage{natbib}
\usepackage{hyperref}
\usepackage{url}

\hypersetup{
  colorlinks=true,
  linkcolor=blue,
  citecolor=blue,
  urlcolor=blue
}

\title{Emergence is Overrated: AGI as an Archipelago of Experts}

\author{Daniel Kilov\\
Machine Intelligence and Normative Theory (MINT) Lab\\
Research School of Social Sciences\\
Australian National University\\
\texttt{daniel.kilov@anu.edu.au}
}

\date{}

\begin{document}

\maketitle

\begin{abstract}
Krakauer, Krakauer, and Mitchell (2025) distinguish between emergent capabilities and emergent intelligence, arguing that true intelligence requires efficient coarse-grained representations enabling diverse problem-solving through analogy and minimal modification. They contend that intelligence means doing ``more with less'' through compression and generalization, contrasting this with ``vast assemblages of diverse calculators'' that merely accumulate specialized capabilities. This paper examines whether their framework accurately characterizes human intelligence and its implications for conceptualizing artificial general intelligence. Drawing on empirical evidence from cognitive science, I demonstrate that human expertise operates primarily through domain-specific pattern accumulation rather than elegant compression. Expert performance appears flexible not through unifying principles but through vast repertoires of specialized responses. Creative breakthroughs themselves may emerge through evolutionary processes of blind variation and selective retention rather than principled analogical reasoning. These findings suggest reconceptualizing AGI as an ``archipelago of experts'': isolated islands of specialized competence without unifying principles or shared representations. If we accept human expertise with its characteristic brittleness as genuine intelligence, then consistency demands recognizing that artificial systems comprising millions of specialized modules could constitute general intelligence despite lacking KKM's emergent intelligence.
\end{abstract}

\section{Introduction}

The rapid proliferation of large language models presents a practical challenge: how do we distinguish genuine progress toward artificial intelligence from mere accumulation of capabilities? With new models released monthly and strong incentives to declare each iteration revolutionary, we need principled criteria for evaluating whether these systems represent steps toward artificial general intelligence or simply more sophisticated calculators.

Krakauer, Krakauer, and Mitchell \citeyearpar{krakauer2025} propose such criteria by examining what makes human intelligence impressive. They identify our apparent ability to wield general representations across disparate domains. The inverse square law, their paradigmatic example, elegantly describes phenomena from gravity to electromagnetism through a single mathematical form. This capacity for ``emergent intelligence,'' doing ``more with less'' through compression and analogical reasoning, represents what we should engineer LLMs to achieve. They explicitly contrast this with systems that merely accumulate domain-specific competencies, warning that ``vast assemblages of diverse calculators'' would not constitute genuine progress toward intelligence.

This framing has immediate implications for AI development. If KKM are correct, engineering efforts focused on adding specialized modules fundamentally misunderstand intelligence. We should instead pursue architectures that discover unifying principles and enable analogical transfer.

This view faces a problem, however; it sets the bar for what counts as intelligence so high as to rule out almost all of intelligent human action. Most human problem-solving looks less like wielding Occam's Razor and more like rummaging through a well-stocked toolbox, applying learned patterns to familiar situations.

This paper advances an alternative conception for what AGI might look like: an archipelago of experts. Rather than awaiting the discovery of some hidden contiguous cognitive continent unified by deep principles, AGI may emerge as a vast collection of specialized islands, each supreme within its domain, connected only by tenuous bridges. If human intelligence consists primarily of brittle, domain-specific expertise, then an artificial system comprising millions of specialized modules should similarly qualify as general intelligence, despite lacking the emergent intelligence that KKM valorize.

The paper proceeds as follows. Section~\ref{sec:emergent} explicates KKM's distinction between emergent capabilities and emergent intelligence. Section~\ref{sec:analogy} demonstrates that analogical reasoning---the cornerstone of their framework---is surprisingly rare and difficult even for humans. Section~\ref{sec:brittleness} presents empirical evidence for the brittleness of human expertise across domains. Section~\ref{sec:archipelago} develops the archipelago thesis and its implications for conceptualizing AGI. The conclusion argues that accepting this reconceptualization fundamentally changes both our expectations for and recognition of artificial general intelligence.

\section{Emergent Intelligence versus Emergent Capabilities}
\label{sec:emergent}

Krakauer, Krakauer, and Mitchell \citeyearpar{krakauer2025} draw a critical distinction between emergent capabilities and emergent intelligence in their analysis of Large Language Models. While emergent capabilities refer to new functional abilities that arise from scaling or reorganization of system components, emergent intelligence represents a fundamentally different phenomenon characterized by the efficient use of coarse-grained representations to solve diverse problems with minimal resources.

The authors define emergent capabilities as task-specific competencies that may appear suddenly with increased scale or through internal reorganization. For instance, they discuss how LLMs demonstrate improved accuracy on benchmarks like three-digit addition when scaled from 6 billion to 175 billion parameters, jumping from 1\% to 80\% accuracy \citep[p.~3]{krakauer2025}. Similarly, the OthelloGPT transformer developed internal board representations despite being trained only on legal move sequences, potentially exhibiting an emergent capability through compression \citep[p.~9]{krakauer2025}. However, the authors emphasize that such capabilities, while impressive, do not constitute intelligence.

In contrast, emergent intelligence involves ``the internal use of coarse-grainings to solve a broad range of problems'' where solutions are ``often related through analogy, that by means of thoughtful and minimal modification, permit significant generalization and extrapolation'' \citep[p.~2]{krakauer2025}. The authors illustrate this with the example of the inverse square law in physics, which through analogical reasoning applies to gravity, electrostatics, acoustics, and electromagnetism. This represents intelligence because it demonstrates ``more with less''---explaining increasing phenomena with a modest set of basic ideas---rather than ``more with more'' \citep[p.~3]{krakauer2025}.

The authors argue that human intelligence exemplifies true emergence through its low-bandwidth efficiency and creation of ``effective theories of knowledge---compact and low-dimensional languages that screen-off most of neuroscience and even much of psychology'' \citep[p.~13]{krakauer2025}. They cite experimental evidence showing humans can configure neural network structures through verbal instruction in minutes, while other primates require laborious trial-and-error training to achieve the same representational geometry \citep[p.~13]{krakauer2025}. This ``instant'' intelligence based on understanding represents a qualitative difference from mere capability accumulation.

Crucially, while LLMs may possess numerous emergent capabilities potentially exceeding human performance in specific domains, the authors contend they have not met the bar for genuine intelligence. LLMs operate more like ``vast assemblages of diverse calculators'' rather than analogy-making systems that can efficiently modify and repurpose concepts \citep[pp.~2--3]{krakauer2025}. The distinction matters because emergent intelligence enables doing ``more with less'' through parsimonious solutions that are readily communicable and modifiable, whereas emergent capabilities may simply represent ``more with more''---increasing functions without the efficiency and generalizability that characterizes true intelligence \citep[pp.~12--13]{krakauer2025}.

KKM's framing resonates with our everyday sense that there's a gap between getting lots of answers right and actually grasping the few deep ideas that make many answers fall out at once. By insisting on compression, reuse, and analogy as the hallmark, they give a clear yardstick for what we should count as progress. It also sharpens practice: it suggests concrete probes (bandwidth use, sample efficiency, cross-task transfer under minimal modification) and encourages building benchmarks that reward unification over accumulation. Even for those of us bullish on today's models, this view doesn't move the goalposts so much as render them visible: it explains why we celebrate inverse-square-law style leaps and how to tell when a system has learned to repurpose an idea rather than just extend a lookup table.

\section{Analogical Reasoning: Analogous to Setting the Bar at Olympic Heights}
\label{sec:analogy}

Although analogical reasoning is a powerful route to generalisation in human reasoning and a hallmark of our species, it is also highly demanding. In fact, we are often surprisingly bad at it. Consider a famous study by \citet{gick1980}, wherein the researchers conducted a series of experiments investigating whether people could use analogical reasoning to solve a classic insight problem requiring participants to determine how to destroy a stomach tumor using radiation without damaging the surrounding healthy tissue. The solution involves converging multiple low-intensity rays from different directions onto the tumor, allowing the rays to summate at the target while remaining harmless to the healthy tissue they pass through individually.

In their experimental paradigm, participants were first presented with military story analogies that embodied structurally similar solutions. The primary analogy involved a general who needed to capture a fortress protected by mined roads that would detonate if crossed by large forces. The general's solution, to divide his army into small groups that converge simultaneously from multiple directions, maps directly onto the radiation problem's dispersion solution.

When participants were not told the military story might be relevant, spontaneous transfer was remarkably poor. In Experiment~IV, when the story was embedded in a memory task, only 20\% of participants (3 of 15) generated the convergence solution, with perhaps only one truly recognizing the analogy unprompted. Experiment~V yielded similar results, with 35--43\% success rates without hints.

In contrast, when explicitly instructed to use the story as a hint, 100\% of participants solved the radiation problem in Experiment~I, and 76\% produced complete or partial solutions in Experiment~II.

Similarly, \citet{novick1991} examined how people transfer mathematical problem-solving procedures via analogy. College students first studied a problem involving determining the right number of plants to buy to pattern a garden bed. The solution involved finding the least common multiple (LCM) of divisors with constant remainders, and they were then asked to solve analogous problems about arranging band members and packaging cookies. The target problems preserved this LCM structure while varying surface features.

Despite explicit hints about numerical correspondences, successful transfer remained surprisingly low. Only 50--68\% succeeded even when told exactly which numbers matched.

Such narrow acquisition and application of abstract knowledge may be the norm, rather than the exception. As \citet{gardner1991} notes, honour grade physics students often struggle to solve basic problems when they're presented differently than the exact format they learned and were tested on. The same is true for students in other disciplines. A study by \citet{voss1986} examined how college graduates answered questions about economic topics. The researchers discovered that participants who had studied economics performed no better than those without economics coursework, prompting them to conclude that the results suggest that classroom instruction in economics does not necessarily lead to superior performance on ``everyday'' economics tasks.

I do not wish to overstate the case; humans can, of course, learn to apply abstract concepts across domains. But when they do, they typically do so via the accumulation of domain specific knowledge. In KKM's terms, they do ``more with more'' by adding to their stores of knowledge.

As I argue in \citet{kilov2021}, reasoning by analogy requires domain-specific knowledge and skills. This is so because domain specific knowledge is required to overcome at least two obstacles to successful analogical reasoning. First, we may lack the subject-specific knowledge needed to recognize what kind of problem we are facing and thus fail to notice that an appropriate analogy is available, as in the stomach-cancer study discussed previously. Recognizing such deep structural similarities depends on substantive knowledge, and in humans, recognitional abilities must be practiced until they are fast enough to be deployed in real time. Second, even when we recognize that an analogy fits, we may still lack the subject knowledge required to apply it correctly, as was the case in the LCM study.

If KKM's conditions for genuine intelligence require that systems demonstrate ``the internal use of coarse-grainings to solve a broad range of problems'' where solutions are ``often related through analogy, that by means of thoughtful and minimal modification, permit significant generalization and extrapolation'' \citep[p.~2]{krakauer2025}, then it follows that much of human cognition fails to qualify as genuine intelligence! We humans typically solve new problems not through elegant compression but through accumulating domain-specific knowledge.

Of course, the examples KKM offer in their paper suggest that maybe the target of their analysis is something more like expert intellectual performance. They cite the inverse square law as paradigmatic of emergent intelligence: a single coarse-grained concept that, through analogical reasoning, elegantly describes gravity, electrostatics, acoustics, and electromagnetism. They quote Poincar\'{e}'s observation that ``Mathematics is the art of giving the same name to different things,'' celebrating the mathematician's ability to do ``more with less'' by finding unifying principles across disparate phenomena. And intuitively, here they seem to be on better ground. After all, at first blush, experts appear to be able to flexibly deploy their skills in contexts that would flummox beginners and this appears to be mediated by a recognition of the deep-structural features of cases. As we shall see in the next section, however, even expert performance is surprisingly brittle and may be much more like the ``vast assemblages of diverse calculators'' that KKM explicitly contrast with genuine intelligence than the ideals captured by their notion of emergent intelligence.

\section{The Brittleness of Expertise}
\label{sec:brittleness}

If everyday expertise involves grasping deep principles that transfer across situations, as KKM's framework suggests, then experts should demonstrate flexible, analogical reasoning within their domains.

However, as \citet{feltovich2006} point out, expertise typically develops in very narrow and highly specific ways, such that there is ``little transfer from high-level proficiency in one domain to proficiency in other domains---even when the domains seem, intuitively, very similar'' (p.~65). For example, surgical expertise is remarkably task-specific. The ability to perform one surgical procedure derives from dedicated practice of that particular task and fails to generalize even to closely related procedures \citep{wanzel2002}. This specificity extends beyond actual surgical tasks: experienced surgeons show little advantage on training exercises explicitly designed to approximate their domain of expertise. \citet{vansickle2007} found no correlation between surgeons' years of practice or number of completed laparoscopic procedures and their performance on a virtual reality laparoscopic simulator---only prior experience with that specific simulator predicted performance. Similarly, \citet{park2007} demonstrated that simulator performance poorly predicted clinical performance. Rather than undermining the value of surgical expertise, these findings illuminate a crucial feature of expert skill: its exquisite sensitivity to the precise contours of a domain.

Could KKM take this evidence to support their distinction? Perhaps surgeons don't transfer between procedures precisely because surgery is capability without intelligence in KKM's sense. After all, it's skilled motor control, not abstract reasoning. I think not, since in earlier work with Jason Stanley, John Krakauer has rightly argued that there is no sharp dividing line between procedural and propositional knowledge \citep{stanley2013}. Both are truly cognitive in the relevant sense.

Besides, such brittleness is not limited to domains that involve high levels of motor skill. Neurologists perform poorly when diagnosing cardiac cases \citep{rikers2002}. \citet{nodine1998} compared search performances of expert radiologists and non-experts on non-radiographic images, such as \emph{Where's Waldo} images. In these tasks, radiologists performed no better than the non-experts; they did not experience a generic boost to search performance. Similar encapsulation also occurs in writing expertise: a technical writer who specializes in instruction pamphlets for home electronics cannot simply pivot to writing newspaper articles \citep{kellogg2018}.

Further, and perhaps more problematically for KKM, is that expertise typically fails to generalise to novel cases \emph{within} an expert's domain. Even small changes can be enough to degrade expert performance. \citet{saariluoma1991}, for example, showed that blindfold-chess experts lost track of piece locations when moves were random rather than conventional. This accords with the classic work of \citet{chase1973}, who found that the recall advantage of chess experts largely disappears when positions are randomized instead of drawn from real games. Expertise can also reduce flexibility in performance: \citet{sternberg1992} compared expert and novice bridge players under arbitrary rule changes and found that the more expert the player, the harder it was to adapt. A similar pattern appears in accounting, where expert accountants struggled more than novices to apply a new tax law that superseded an old one \citep{marchant1991}. \citet{kilov2021} discusses chess problem-solving studies \citep{bilalic2008, saariluoma1990} in which a series of puzzles shared an initial motif: after recognizing the pattern in the first four, experts reused it on a fifth puzzle and overlooked an objectively better solution, whereas less experienced players---less affected by the Einstellung stimulus---found the superior approach.

Why is expertise so often brittle? One reason is that as task or situational demands increase, the range of acceptable responses shrinks. Playing a game of chess against me, Magnus Carlsen could use literally any opening he liked and still win. This would not be true if he were playing against someone of similar calibre. Consequently, the representational structures that mediate expert performance must become increasingly specialized. To borrow an analogy from \citet{kilov2021}, pocket-knives are general-purpose tools. They come in handy across a wide range of situations. While you can certainly use a pocketknife to uncork a bottle of wine, it will never match a dedicated corkscrew. This is why so many pocketknives include corkscrew attachments. Yet the corkscrew's specialized design renders it useless for nearly any other task. The very features that make it excel at one function explain why it fails at most others. This suggests that much of expert reasoning simply does not involve the kind of domain-general reasoning of the kind KKM want to identify with emergent intelligence.

I acknowledge that my account leaves unresolved a question that KKM's framework intuitively answers: if human expertise is as brittle and domain-specific as I have argued, why does it so often \emph{appear} flexible and generalizable? The answer, I suggest, involves distinguishing genuine flexibility from its simulacrum. In some cases, experts do demonstrate authentic cross-domain transfer of the kind KKM celebrate; their examples of the inverse square law and similar scientific unifications are genuine instances of emergent intelligence. More commonly, however, what appears to be flexible expertise is actually the accumulated result of extensive pattern learning across numerous specific contexts. The expert has simply encountered so many variations that situations appearing novel to observers are in fact familiar territory, each handled by its own encapsulated routine rather than by any unifying principle. This accumulation of individually inflexible patterns creates an illusion of generality. The expert does indeed handle diverse situations successfully, but through ``more with more'' rather than ``more with less.''

Consider the revealing case of cyborg chess, which offers a natural experiment in decomposing expert intelligence. When Gary Kasparov organized the first human-computer team tournament in 1998, the results challenged conventional assumptions about the nature of chess mastery. As \citet{epstein2021} observes, the computer partnership ``changed the pecking order instantly'' by handling tactical calculations, allowing humans to focus purely on strategy. Kasparov, who had defeated one opponent 4--0 in traditional play just a month earlier, could only manage a 3--3 draw in the cyborg format. His own assessment was stark: his advantage in calculating tactics had been nullified by the machine \citep[p.~49]{epstein2021}.

This dramatic leveling reveals that Kasparov's dominance rested primarily on pattern recognition and rapid calculation---precisely what could be outsourced to silicon. Yet it would be absurd to conclude that his mastery, built on recognizing thousands of positions and accessing vast game knowledge, does not constitute intelligent reasoning. If KKM's framework excludes such pattern-based expertise as mere ``database lookup'' rather than genuine intelligence, it risks defining away many of humanity's most impressive cognitive achievements.

If what I've said is true, however, it appears to make a mystery of the creative abilities of experts. How, for example, do expert scientists reliably generate novel discoveries? How do great artists produce strings of great works? The answer, surprisingly, is that they don't. Decades of research instead suggest that creative breakthroughs are the product of evolutionary processes.

The theory of Blind Variation and Selective Retention (BVSR) posits that creativity proceeds via the generation of multiple, partly ``blind'' candidate ideas---produced without reliable foresight of their value---followed by the selective retention and further development of the few that prove useful, original, or surprising \citep{campbell1960, simonton2010, simonton2013, simonton2022}. A key quantitative implication is the equal-odds rule: the probability that any given product is a ``hit'' is roughly constant across attempts, so producing more work increases the absolute number of successes even if average quality per attempt does not---a pattern observed across individuals and domains \citep{simonton2010, jung2015}. Large-scale ``science of science'' analyses converge: the impact of a scientist's most influential paper is randomly distributed within their career sequence, consistent with BVSR's stochastic generation and subsequent selection \citep{sinatra2016}. Together, these lines of evidence support a view of creative achievement as a probabilistic search---often guided but never fully prescient---in which output volume (variation) and evaluative filtering (selection) jointly determine realized breakthroughs \citep{simonton2022}.

The cases that KKM celebrate, of scientists who seem to wield unifying principles to generate breakthrough after breakthrough, may simply represent the lucky tail of a probability distribution. Consider a parallel: if ten thousand people each flip a fair coin twenty times, approximately ten will get fifteen or more heads purely by chance. To an observer focused only on these exceptional cases, it might appear that these individuals possess a special coin-flipping technique or deeper understanding of probability. Similarly, when thousands of scientists attempt to find unifying principles across domains, a few will succeed multiple times through fortunate conjunctions of preparation and opportunity.

\section{AGI as an Archipelago of Experts}
\label{sec:archipelago}

The analysis presented in this paper raises questions about what should constitute artificial general intelligence. If human intelligence consists primarily of brittle, domain-specific expertise rather than the elegant analogical reasoning that KKM identify with emergent intelligence, then we must reconsider our conceptual framework for AGI.

I propose that an archipelago of specialized experts would constitute genuine general intelligence, despite failing to meet KKM's criteria for emergent intelligence. This claim follows directly from the empirical evidence about human expertise. As documented throughout this paper, human experts achieve remarkable performance through massive accumulation of domain-specific patterns rather than through compression and analogical reasoning. Kasparov's chess mastery, surgical expertise, and other paradigmatic cases of human intelligence operate through ``more with more'' rather than ``more with less.'' If we accept these as manifestations of intelligence, consistency demands that we extend the same recognition to artificial systems with similar architectures.

The archipelago metaphor captures the essential structure of such intelligence: isolated islands of extreme competence without unifying principles or shared representational schemes. Each module would excel within its territory through exhaustive pattern recognition and specialized heuristics. Such a system would lack the characteristics of emergent intelligence as KKM define it. There would be no elegant compression, no powerful analogies spanning domains, no parsimonious theories that explain vast phenomena with minimal concepts.

Yet this archipelago would possess capabilities far exceeding human performance across countless domains. Consider a system comprising millions of specialized modules: one achieving general human diagnosis of retinal diseases, another optimizing supply chains, another parsing legal documents, each supreme within its narrow scope. The collective capability of such a system would be substantial, even if no single module or organizing principle could be identified as demonstrating emergent intelligence in KKM's sense.

This reconceptualization has implications for how we think about the path to AGI. Rather than waiting for systems that demonstrate true emergent intelligence through compression and analogy, we might already be witnessing the emergence of AGI through the gradual accumulation of specialized capabilities. Current large language models, despite their limitations, already function as vast assemblages of specialized patterns and heuristics. Their intelligence, such as it is, resembles the ``database expertise'' of human experts more than the theoretical elegance KKM valorize.

The brittleness and lack of transfer that characterize both human expertise and current AI systems should not be seen as failures to achieve ``true'' intelligence. Instead, they reflect the nature of competence in complex domains where details matter and problems resist elegant compression. An archipelago of experts represents a different form of intelligence than KKM envision, but it is intelligence, nonetheless and certainly an intelligence worth having. To deny this would be to exclude most human achievement from the category of intelligence, a conclusion that seems both counterintuitive and unhelpful for understanding the forms that artificial general intelligence might actually take.

\section{Conclusion}

If human expertise operates through brittle, domain-specific pattern recognition rather than elegant compression, then an archipelago of specialized AI experts would constitute genuine general intelligence, even without KKM's unifying principles. AGI may emerge not through breakthroughs in analogical reasoning, but through the mundane accumulation of narrow capabilities. The path to artificial general intelligence might already be underway, one isolated competency at a time.

Perhaps we've been searching for the cognitive mainland when intelligence was always an archipelago.

\bibliographystyle{apalike}
\bibliography{references}

\end{document}